\documentclass[letterpaper, 10 pt, conference]{ieeeconf}  

\IEEEoverridecommandlockouts                              

\usepackage{multicol}
\usepackage{graphicx}
\usepackage[font=small,labelfont=bf,justification=justified]{caption}
\usepackage{subcaption}
\usepackage{amsmath}
\usepackage{amssymb}
\usepackage{mathtools}
\usepackage{textcomp}
\usepackage{gensymb}
\usepackage{bm}
\usepackage{float}
\usepackage{color}
\usepackage{multirow}
\usepackage{hhline}
\usepackage{changepage}
\usepackage{comment}
\usepackage{titlesec}
\titlespacing\section{0pt}{3pt}{3pt}

\usepackage{cite}

\usepackage[ruled,vlined,linesnumbered]{algorithm2e}

\usepackage[hidelinks, bookmarks=true]{hyperref}

\newcommand{\nonl}{\renewcommand{\nl}{\let\nl\oldnl}}

\title{\LARGE \bf
A General Formulation for Path Constrained Time-Optimized Trajectory Planning with Environmental and Object Contacts
}
\author{Dasharadhan Mahalingam$^1$, Aditya Patankar$^1$, Riddhiman Laha$^2$, Srinivasan Lakshminarayanan$^2$, \\Sami Haddadin$^2$, and Nilanjan Chakraborty$^1$
\thanks{$^{1}$The authors are with the Department of Mechanical Engineering, 
        Stony Brook University, USA.
       {\texttt{\{dasharadhan.mahalingam, aditya.patankar, nilanjan.chakraborty\}@stonybrook.edu.}}}%
\thanks{$^{2}$The authors are with Munich Institute of Robotics and Machine Intelligence (MIRMI), Technical University of Munich. Email:\texttt{\{riddhiman.laha,sami.haddadin\}@tum.de}}%
\thanks{This work is partially supported by the US Department of Defense through  ALSRP under Award No. HT94252410098.}
}
\renewcommand{\baselinestretch}{0.93}
\begin{document}

\maketitle
\thispagestyle{empty}
\pagestyle{empty}

\begin{abstract}
A typical manipulation task consists of a manipulator equipped with a gripper to grasp and move an object with constraints on the motion of the hand-held object, which may be due to the nature of the task itself or from object-environment contacts. In this paper, we study the problem of computing joint torques and grasping forces for time-optimal motion of an object, while ensuring that the grasp is not lost and any constraints on the motion of the object, either due to dynamics, environment contact, or no-slip requirements, are also satisfied. We present a second-order cone program (SOCP) formulation of the time-optimal trajectory planning problem that considers nonlinear friction cone constraints at the hand-object and object-environment contacts. Since SOCPs are convex optimization problems that can be solved optimally in polynomial time using interior point methods, we can solve the trajectory optimization problem efficiently. We present simulation results on three examples, including a non-prehensile manipulation task, which shows the generality and effectiveness of our approach. 


\textit{Video}--- \url{https://youtu.be/8PLYkxJ3TJU}
\end{abstract}

\section{Introduction}\label{sec:intro}
 Time-optimized motion planning promises attractive new applications that will transform industrial productivity as well as improve service robotics utilization~\cite{zhou2022review}. For example, an optimized motion planner could deliver fragile objects on a factory floor within a stipulated time such that the objects do not fall over. In a service robotics setting, the same planner can be employed to transport a bowl of soup or a cup of water to a person with reduced mobility~\cite{laha2022coordinate, laha2023predictive}. As far as robot manipulators are concerned, completely solving this planning problem is extremely complex and challenging~\cite{lin1983formulation,bobrow1985time,shiller1989robot}. The problem becomes even harder when environmental and object contacts have to be taken into account~\cite{suomalainen2022survey}. 

A two-step approach is usually used for time-optimal trajectory planning~\cite{verscheure2008practical,laha2023s}. In the first step, a collision-free path is planned, and in the second step, the dynamics of the system, as well as the actuator limits, are taken into account to convert the path to a minimum-time trajectory. 
In this paper, we focus on the second step for time-optimal object manipulation while considering contact constraints along with dynamic constraints, actuator limits, and any path constraints that might be present.
\begin{figure}[t]
    \begin{center}\includegraphics[width=\linewidth]{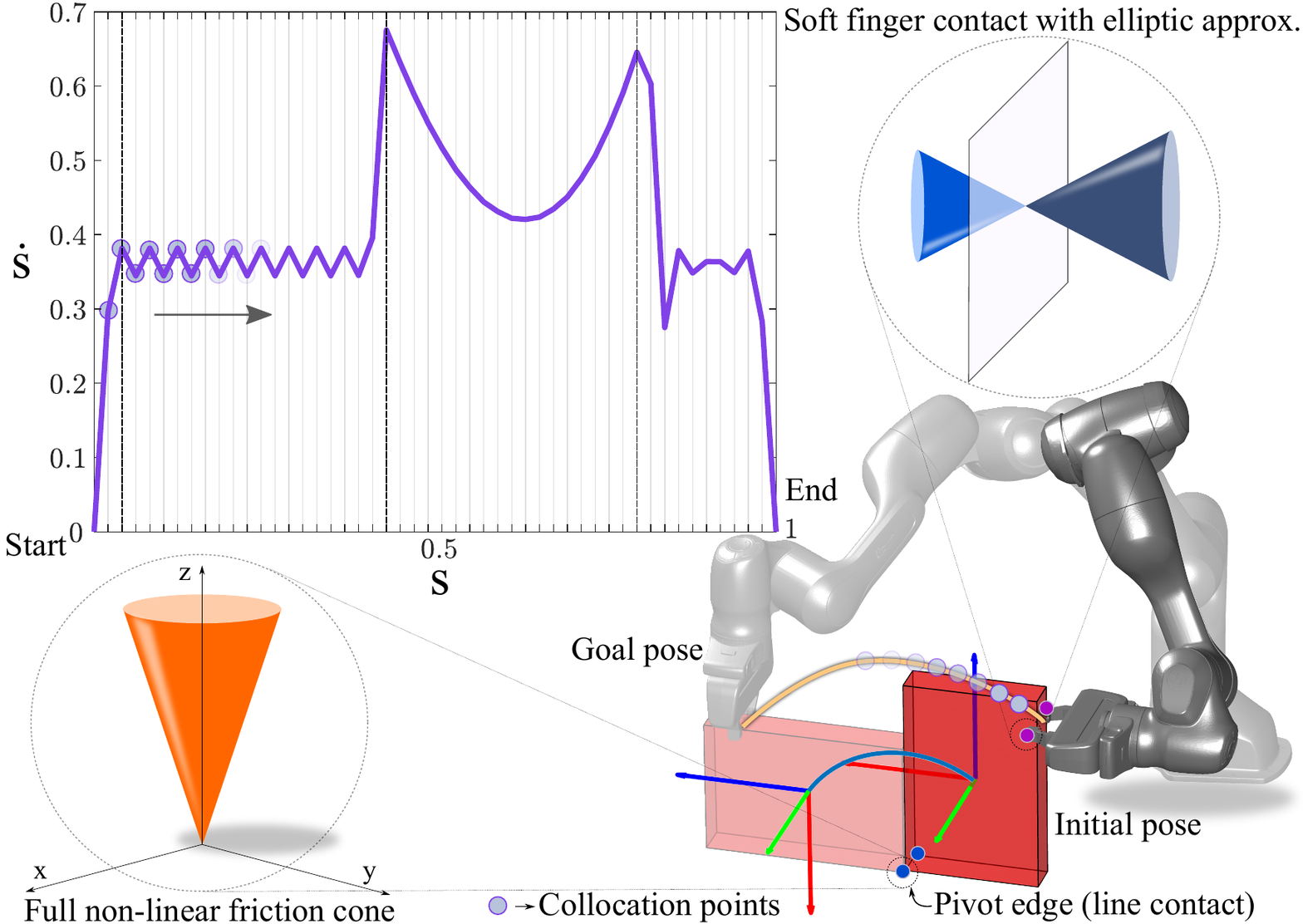}
    \caption{%
   Example pivoting task, which is a path constrained robot task involving manipulator object and object environment contacts. Our method takes into account all the kinematic and dynamic constraints of the system to compute the time-optimal trajectory. The $(s,\dot{s})$ phase plane, which connects the initial path position to the final position, illustrates our problem. The purple curve is the time scaling that we achieve while respecting position, velocity, and acceleration constraints. The blue dots denote the maximum feasible scaling possible at the collocation points. The vertical dashed lines represent specific instances when joints $6,7,$ and $4$ are about to hit the limits. Note that we consider the full non-linear friction cone constraints and the soft finger contact with elliptic approximations for modeling the two types of contacts.   
}
    \label{fig:cover_pic}
    \end{center}
    \vspace{-25pt}
\end{figure}
 
 Contact with the environment during motion planning introduces additional complexities due to the underlying frictional dynamics~\cite{kao2016contact}. 
 However, contacts with the environment can be exploited to aid manipulation~\cite{patankar2020handobject, fakhari2021motion}. For example, a \textit{heavy} object can be manipulated by pivoting instead of lifting (Fig~\ref{fig:cover_pic}). Reaction forces generated at the environment-object contact can be used to partly balance the weight of the object. 
 Therefore, our goal is to take contact constraints into account for trajectory optimization. To this end, we consider the complete nonlinear friction cone constraints instead of a polygonal approximation~\cite{lynch2017modern,selvaggio2023non}. 
 This leads us to the problem we aim to focus: \textit{How to bridge the gap between geometric motion planning and time-optimal trajectory planning such that the robot-object and environment-object contact constraints of the system are respected?}

{\bf Contributions}: We present a novel second-order cone program (SOCP) formulation of the time-optimal trajectory planning problem that considers nonlinear friction cone constraints at the hand-object and object-environment contacts along with the constraints imposed by the robot dynamics and the robots' actuator limits. Our model is general enough to handle multiple manipulators making contact with an object as well as multiple contacts between the object and the environment. Since SOCPs are convex optimization problems that can be solved optimally in polynomial time using interior point methods, we can solve the trajectory optimization problem efficiently. We present simulation results on three examples, including a non-prehensile manipulation task, which shows the generality and effectiveness of our approach.

\section{Related Work}\label{sec:rel_work}

The time-optimal trajectory planning problem for robotic manipulators has been a significant area of research in the literature. The core objective is to determine the optimal timing of motion along a predefined path, subject to various constraints, such as actuator torque limits and dynamic considerations. Traditionally, robot trajectory generation can be categorized into (a) offline and (b) online (real-time) strategies. Offline trajectory generation has a long history starting from the late \textquotesingle$60$s. Some of the seminal works in this regard are the solutions to minimum-time control problems~\cite{kahn1969near,kahn1971near}. However, including collision-free trajectories in the introduced formulation is non-trivial. 

The authors in~\cite{niv1984optimal} overcame this limitation by proposing a parameter optimization scheme that guaranteed minimum time travel using the maximum available torques while avoiding obstacles in the workspace. Similar directions were also explored in~\cite{hollerbach1983dynamic}, where scaling was performed on existing path velocity profiles to maximize the output of the actuators. Similar algorithms were proposed in~\cite{bobrow1985time,shin1985minimum}, which exploited switching curves in the phase plane for end-effector motion along a path. The assumption for the latter one is that actuator torque bounds are a quadratic function of the joint velocity, with joint positions being parameterized by a polynomial path parameter. Recent approaches, in contrast, focus more on time-optimized path tracking, considering velocity, acceleration, and jerk limits~\cite{kunz2013time,pham2014general,lange2015path}. Our approach, although similar to~\cite{verscheure2009trajopt}, differs considerably in the sense that we extend it to a more general scheme. Regardless, most of the works do not consider held objects and model explicitly the robot-object contact that we take care of during the motion execution. On the other hand, on-line paradigms fuse the path and time parameterization as a single step~\cite{kroger2009online}. Typically, offline methods could be able to generate more optimal trajectories, although they are not suited for real-time applications owing to their computational load. In contrast, online methods are efficient, yet prone to discontinuities in the commanded trajectory \cite{he2022online}. 

The problem of computing optimal contact forces, at the manipulator-object contacts, was first formulated as a Second-Order Cone Program (SOCP) by~\cite{lobo1998applications}. A fast interior-point algorithm for solving this formulation was proposed in~\cite{boyd2007contact} with the point contact with friction model as the cone constraint. Building on this, the authors in~\cite{patankar2020handobject} presented a SOCP for computing the optimal contact forces, while considering contact with the environment, for \textit{quasi-statically} manipulating heavy objects by pivoting. The proposed formulation is general enough to incorporate manipulator joint limit constraints and has been integrated with a task space-based planning approach for pivoting~\cite{fakhari2021motion}. In this work, we build upon the formulation proposed in~\cite{patankar2020handobject} while considering the dynamics of the robot as well as the inertial properties of the object(s) being manipulated.   

The authors in~\cite{sleiman2019contact} formulate a \textit{contact-implicit} trajectory optimization problem, as a nonlinear program, for the task of pushing an object along a plane surface to a desired goal pose. The contact constraints have been represented as linear complementarity constraints which approximate the friction cones. Similar approaches~\cite{onol2018comparative, onol2019contact, onol2020tuning} have been developed for tasks such as planar pushing in which an optimal sequence of control inputs has to be provided to make contact with the object and apply the necessary normal contact forces to push the object further away from the robot. In contrast, we are more interested in manipulation tasks which require the robot to maintain continuous contact, at more than one location, with the object throughout the motion.

\section{System Modeling And Assumptions}\label{sec:sys_mod}
We consider $v$ robot manipulators manipulating an object. Let there be $u$ object-environment contacts.
Coordinate frames $\{E_i\}$ (where $i=1,2,\dots,u$) and $\{M_j\}$ (where $j=1,2,\dots,v$) are fixed at each environment and manipulator contact respectively such that the $\bm{z}$-axis of the coordinate frames are normal (inward) to the object surface and the other two axes, $\bm{x}$ and $\bm{y}$ are tangent to the surface. We assume that a collision-free geometric path is available as some waypoints for the robot manipulator which we want to convert to a time-optimal trajectory. 

\subsection{Manipulator Kinematics}
Using the product of exponentials (POE) formula~\cite{lynch2017modern}, the forward kinematics of an $n$ DOF serial manipulator is
\begin{equation}
    \label{eq:manip_fk}
    \mathbf{X} = e^{\widehat{\bm{\xi}}_1q_1}e^{\widehat{\bm{\xi}}_2q_2}\dots~e^{\widehat{\bm{\xi}}_nq_n}\mathbf{X}_{ref},
\end{equation}
where $\bm{\xi}_l \in se(3), l = 1,2,\dots,n$ are the twists contributed by each joint and $\mathbf{X}_{ref} \in SE(3)$ is the reference configuration of the manipulator at $\bm{\Theta} = \bm{0}$.
The body velocity of the end-effector of the manipulator, $^X\mathbf{V} \in \mathbb{R}^6$ is related to its joint rates through the body Jacobian $\mathbf{J}^b(\bm{q}) \in \mathbb{R}^{6\times n}$ as
\begin{equation}
    \label{eq:manip_vel_kin}
    ^X\mathbf{V} =
    \begin{bmatrix}
        \bm{v}^b \\ \bm{\omega}^b
    \end{bmatrix} = 
    \mathbf{J}^b(\bm{q})\dot{\bm{q}} = 
    \begin{bmatrix}
        \mathbf{J}^b_v(\bm{q}) \\ \mathbf{J}^b_\omega(\bm{q})
    \end{bmatrix}\dot{\bm{q}},
\end{equation}
where $\bm{v}$ and $\bm{\omega}$ denote the linear and angular velocities respectively, $\bm{q}$ is the vector of joint displacements and $\dot{\bm{q}}$ is the vector of joint rates. The body acceleration of the end-effector of the manipulator can be determined from
\begin{equation}
    \label{eq:manip_acc}
    {^X\dot{\mathbf{V}}}  = 
    \begin{bmatrix}
        \dot{\bm{v}}^b \\ \dot{\bm{\omega}}^b
    \end{bmatrix} = 
    \mathbf{J}^b(\bm{q})\ddot{\bm{q}} + \dot{\mathbf{J}}^b(\bm{q})\dot{\bm{q}}.
\end{equation}

\subsection{Manipulator Dynamics}
Extending our analysis to the acceleration level, the equations of motion for the robot manipulator in contact with the environment can be written as,
\begin{align}
    \label{eq:manipulator_eom}
    \bm{\tau} - \mathbf{J}^T(\bm{q})\bm{h}_e = \mathbf{M}(\bm{q})\Ddot{\bm{q}} + \mathbf{C}(\bm{q},\dot{\bm{q}})\dot{\bm{q}} + \mathbf{g}(\bm{q}),
\end{align}
where $\bm{\tau} \in \mathbb{R}^n$ is the vector of applied joint torques, $\bm{h}_e \in \mathbb{R}^6$ is the wrench exerted on the manipulator by the environment, $\mathbf{J}(\bm{q}) \in \mathbb{R}^{6\times n}$ is the manipulator (geometric) Jacobian matrix, $\mathbf{M}(\bm{q}) \in \mathbb{R}^{n\times n}$ is the positive definite mass matrix, $\mathbf{C}(\bm{q},\dot{\bm{q}}) \in \mathbb{R}^{n\times n}$ is the matrix accounting for Coriolis and centrifugal effects and $\mathbf{g}(\bm{q}) \in \mathbb{R}^n$ is the gravity term.

We are primarily interested in computing the joint torques required to produce the necessary wrenches at the manipulator-object contacts. Therefore, we need to establish a relationship between the joint torques and contact wrenches, $\bm{F}_M$ exerted on the object. Let $n_j$ be the number of DOFs of the $j$-th manipulator, $\bm{q}_j = [q_{j,1}, q_{j,2},\dots, q_{j,n_j}]^T$ be the joint positions and $\bm{\tau}_j = [\tau_{j,1}, \tau_{j,2},\dots, \tau_{j,n_j}]^T$ be the joint torques of this manipulator. The equations of motion of the $j$-th manipulator can then be formulated as,
\begin{align*}
    \label{eq:jth_manipulator_eom}
    \bm{\tau}_j + \mathbf{J}_j^T(\bm{q}_j)\bm{f}_{M_j} = \mathbf{M}_j(\bm{q}_j)\Ddot{\bm{q}_j} + \mathbf{C}_j(\bm{q}_j,\dot{\bm{q}_j})\dot{\bm{q}_j} + \mathbf{g}_j(\bm{q}_j),
\end{align*}
where $\mathbf{J}_j \in \mathbb{R}^{6\times n_j}$ is the Jacobian matrix of the $j$-th manipulator expressed in the contact frame $\{M\}_j$ and $\mathbf{M}_j(\bm{q}_j), \mathbf{C}_j(\bm{q}_j,\dot{\bm{q}_j}), \mathbf{g}_j(\bm{q}_j)$ are the respective mass matrix, Coriolis and centrifugal effects, gravity term of the $j$-th manipulator. Here, we note that the wrench exerted on the manipulator is the reaction wrench $-\bm{f}_{M_j}$, and hence the $-\mathbf{J}_j^T(\bm{q}_j)(-\bm{f}_{M_j}) = +\mathbf{J}_j^T(\bm{q}_j)\bm{f}_{M_j}$ term on the left side of the equation. By defining $\bm{q} \in \mathbb{R}^n$ and  $\bm{\tau} \in \mathbb{R}^n$ ($n=\sum_{j=1}^{v}n_j$) as $\bm{q} = [\bm{q}_1,\bm{q}_2,\hdots,\bm{q}_v]^{T}$ and $\bm{\tau} = [\bm{\tau}_1,\bm{\tau}_2,\hdots,\bm{\tau}_v]^{T}$,
the relationship between the joint torques of all the manipulators ($\bm{\tau}$) and all the manipulator-object contact wrenches ($\bm{F}_M$) can be represented in compact form as
\begin{equation}
    \label{eq:eom_of_m_manipulators}
    \bm{\tau} + \mathcal{J}_M(\bm{q})\bm{F}_M = \mathcal{M}_M(\bm{q})\Ddot{\bm{q}} + \mathcal{C}(\bm{q},\dot{\bm{q}})\dot{\bm{q}} + \mathcal{G}(\bm{q}).
\end{equation}
where $\mathcal{J}_M = \text{diag}(\mathbf{J}^{T}_1, \mathbf{J}^{T}_2,\dots, \mathbf{J}^{T}_v) \in \mathbb{R}^{n \times 6v}$ is the overall Jacobian matrix of all the manipulators, $\mathcal{M}_M = \text{diag}(\mathbf{M}_1, \mathbf{M}_2,\dots, \mathbf{M}_v) \in \mathbb{R}^{n \times n}$ is the overall mass matrix of all the manipulators, $\mathcal{C}_M = \text{diag}(\mathbf{C}_1, \mathbf{C}_2,\dots, \mathbf{C}_v) \in \mathbb{R}^{n \times n}$ is the overall Coriolis and centrifugal effect of all the manipulators and $\mathcal{G}_M = [\mathbf{g}_1, \mathbf{g}_2,\dots, \mathbf{g}_v]^T \in \mathbb{R}^{n}$ is the overall gravity term of all the manipulators.
\subsection{Object Dynamics}
Let $^O\mathbf{R}_{E_i} \in SO(3)$ and $^O\mathbf{R}_{M_j} \in SO(3)$ be $3\times3$ orthogonal matrices representing the orientation of the \textit{environment-object contact} coordinate frame and \textit{manipulator-object contact} coordinate frame respectively expressed in the object coordinate frame $\{O\}$. Each contact wrench, $\bm{f}_{E_i}\text{ and }\bm{f}_{M_j}$, can be expressed in the object coordinate frame $\{O\}$ as
\begin{align}
    ^O\bm{f}_{E_i} = \mathbf{G}_{E_i}\bm{f}_{E_i} =
    \begin{bmatrix}
        ^O\mathbf{R}_{E_i} & \mathbf{0} \\
        \mathbf{S}({^O\bm{p}_{E_i}}) {^O\mathbf{R}_{E_i}} & ^O\mathbf{R}_{E_i}
    \end{bmatrix}
    \bm{f}_{E_i},
\end{align}
\begin{align}
    ^O\bm{f}_{M_j} = \mathbf{G}_{M_j}\bm{f}_{M_j} =
    \begin{bmatrix}
        ^O\mathbf{R}_{M_j} & \mathbf{0} \\
        \mathbf{S}({^O\bm{p}_{M_j}}) {^O\mathbf{R}_{M_j}} & ^O\mathbf{R}_{M_j}
    \end{bmatrix}
    \bm{f}_{M_j}
\end{align}
Here, $\mathbf{S(.)} \in skew(3)$ is the skew-symmetric operator~\cite{murray2017mathematical}, such that, for any $\bm{a}\in\mathbb{R}^3, \bm{b}\in\mathbb{R}^3, ~\mathbf{S}(\bm{a})\bm{b} = \bm{a}\times\bm{b}$. Also, $^O\bm{p}_{E_i}$ and $^O\bm{p}_{M_j}$ are the positions of $\{E\}_i$ and $\{M\}_j$ expressed in $\{O\}$ respectively.
Therefore, 
the object motion satisfies the equation,
\begin{align}
    \label{eq:object_eom}
    \sum_{i=1}^{u}\mathbf{G}_{E_i}\bm{f}_{E_i} +
    \sum_{j=1}^{v}\mathbf{G}_{M_j}\bm{f}_{M_j} +
    {^O\bm{f}_{ext}} = {^{O}\bm{f}_{net}}.
\end{align}
where $^O\bm{f}_{ext} \in \mathbb{R}^6$ is the total external (force and moment) wrench acting on the object (including the object weight) expressed in ${O}$ and $^O\bm{f}_{net}$ is the net wrench (resultant of all wrenches) acting on the object expressed in ${O}$.

By introducing $\bm{F}_E \in \mathbb{R}^{6u}$ and $\bm{F}_M \in \mathbb{R}^{6v}$ as $\bm{F}_E = [\bm{f}_{E_1},\bm{f}_{E_2},\hdots,\bm{f}_{E_u}]^{T}$ and $\bm{F}_M = [\bm{f}_{M_1},\bm{f}_{M_2},\hdots,\bm{f}_{M_u}]^{T}$
equation (\ref{eq:object_eom}) can be represented in a more compactly as,
\begin{equation}
    \label{eq:object_eom_compact}
    \mathbf{G}_E\bm{F}_E + \mathbf{G}_M\bm{F}_M + {^O\bm{f}_{ext}} = {^{O}\bm{f}_{net}}
\end{equation}
where $\mathbf{G}_E = [\mathbf{G}_{E_1}, \mathbf{G}_{E_2}, \dots, \mathbf{G}_{E_u}] \in \mathbb{R}^{6\times6u}$ is the environment contact matrix and $\mathbf{G}_M = [\mathbf{G}_{M_1}, \mathbf{G}_{M_2}, \dots, \mathbf{G}_{M_v}] \in \mathbb{R}^{6\times6v}$ is the manipulator contact matrix, which is also known as the \textit{grasp map} or \textit{grasp matrix}~\cite{murray2017mathematical}.

\begin{figure}[t]
    \begin{center}\includegraphics[width=0.95\linewidth]{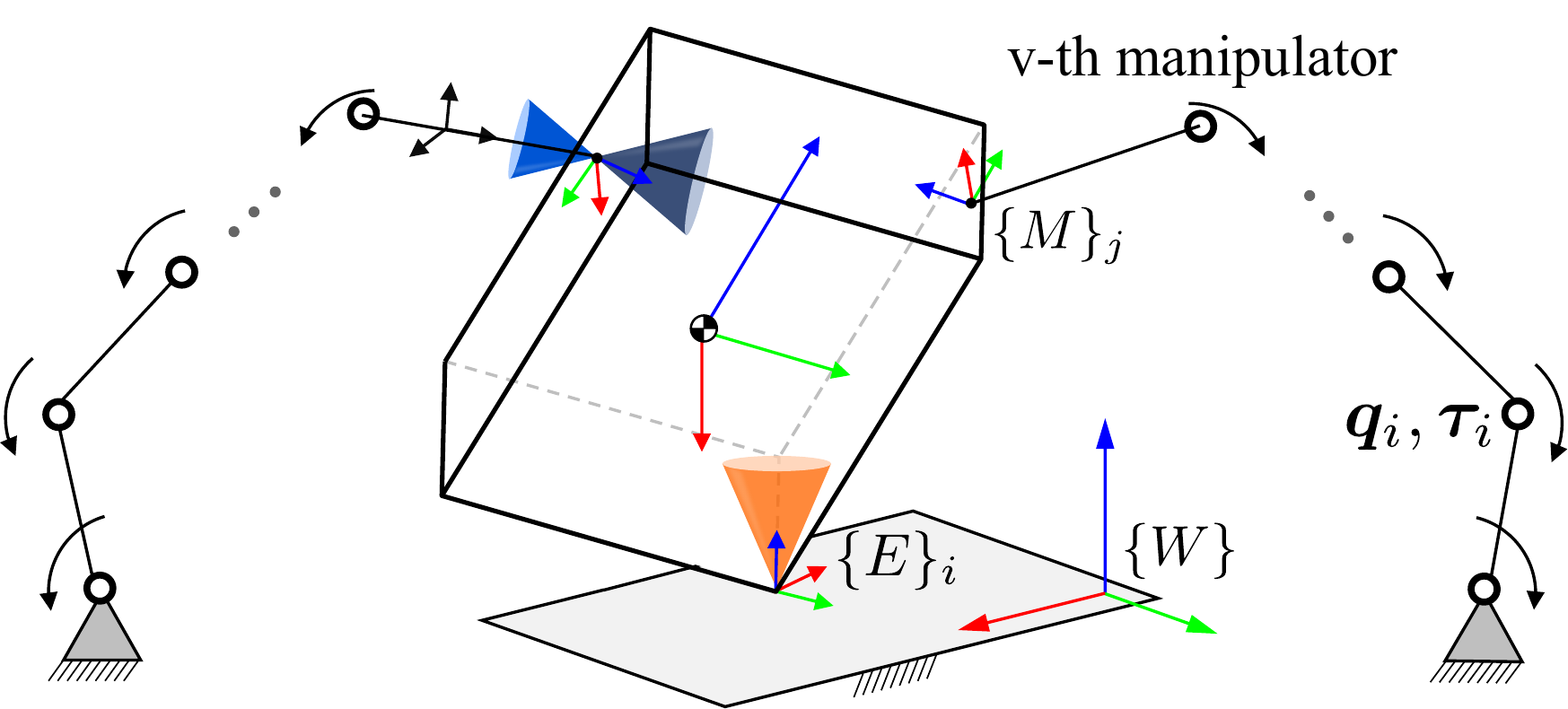}
    \caption{A rigid body being manipulated with $v$ manipulators and in contact with the environment at $u$ points. The contacts between object and the environment are modeled as PCWF and contacts between manipulators and the objects are assumed to be SFCE. $\{E\}$, $\{M\}$, and $\{W\}$ represent the environment, manipulator, and world reference frames.  
}
    \label{fig:fbd_pic}
    \end{center}
    \vspace{-25pt}
\end{figure}

\subsection{Contact Constraints}
We use the point contact with friction (PCWF) for the object-environment contacts, and the soft finger contact with elliptic approximation (SFCE) for the contacts between the manipulators and the object~\cite{patankar2020handobject}. For PCWF, the contact wrench satisfies the friction cone constraint as
\begin{align}
    \label{eq:pcwf_constraint}
    \frac{1}{\mu_{E_i}}\sqrt{\bigg(\frac{f_{E_{i,x}}}{e_{E_{i,x}}}\bigg)^2+
    \bigg(\frac{f_{E_{i,y}}}{e_{E_{i,y}}}\bigg)^2} \leq f_{E_{i,z}},
\end{align}
where the parameters $\mu_{E_i}, e_{E_{i,x}} \text{ and } e_{E_{i,y}}$ are positive constants defining the friction coefficients at the $i$-th environment contact point and $i=1,2,\dots,u$. Similarly, in SFCE, the contact wrench  satisfies the elliptic constraint as
\begin{align}
    \label{eq:sfce_constraint}
    \frac{1}{\mu_{M_j}}\sqrt{\bigg(\frac{f_{M_{j,x}}}{e_{M_{j,x}}}\bigg)^2+
    \bigg(\frac{f_{M_{j,y}}}{e_{M_{j,y}}}\bigg)^2+\bigg(\frac{\tau_{M_{j,z}}}{e_{M_{j,z}}}\bigg)^2}
    \leq f_{M_{j,z}},
\end{align}
where the parameters $\mu_{M_j}, e_{M_{j,x}}, e_{M_{j,y}}, e_{M_{j,z}}$ are positive constants defining the friction coefficients at the $j$-th manipulator contact and $j = 1,2,\dots,v$. The constraints (\ref{eq:pcwf_constraint}) and (\ref{eq:sfce_constraint}) are \textit{Second-Order Cone} (SOC) constraints \cite{boyd2004convex}. For any $\bm{r} \in \mathbb{R}^6$, we define the friction cones $\mathcal{K}_{E_i}$ and $\mathcal{K}_{M_j}$ as
\begin{align*}
    \mathcal{K}_{E_i} = \Biggl\{\bm{r}\Bigg|
    \frac{1}{\mu_{E_i}}\sqrt{\frac{r_1^2}{e_{E_{i,x}}^2}+
    \frac{r_2^2}{e_{E_{i,y}}^2}} \leq r_3, r_{4,5,6}=0
    \Biggr\},
\end{align*}
\begin{align*}
    \mathcal{K}_{M_j} = \Biggl\{\bm{r}\Bigg|
    \frac{1}{\mu_{M_j}}\sqrt{\frac{r_1^2}{e_{M_{j,x}}^2}+
    \frac{r_2^2}{e_{M_{j,y}}^2}+\frac{r_6^2}{e_{M_{j,z}}^2}}
    \leq r_3, r_{4,5} = 0
    \Biggr\},
\end{align*}
where $r_k, k = 1,2,\dots,6$ is the $k$-th component of $\bm{r}$. To ensure the manipulated object is not damaged by the internal forces, we consider an upper bound $\overline{\bm{F}}_{z}$ on the normal contact force. Thus, the normal contact force $\bm{F}_z=[f_{M_{1,z}}, f_{M_{2,z}},\dots, f_{M_{v,z}}]^T \leq \overline{\bm{F}}_{z}\footnote{Note that a set of contact forces that result in no net force on object is termed as \textit{internal forces}.}.$

\section{Mathematical Problem Formulation}\label{sec:prob_def}

\subsection{Time Parameterization of Motion}
Let us assume that the object frame $\{O\}$ goes through a trajectory which can be expressed using a scalar path coordinate $s$, as $\{O\}(s)$. Assuming that the object starts moving at time, $t=0$ and finishes moving through the trajectory at time, $t=T$, the time dependence of the object trajectory can be obtained through the time scaling $s(t)$ which assigns a value to $s$ for any instant of time $t \in [0, T]$, such that $s:[0, T] \mapsto [0, 1]$ with $s(0) = 0$ and $s(T) = 1$. For notational convenience, let us omit the time dependence of the path coordinate $s$ and its derivatives.

From the motion of $\{O\}$, we can then determine the joint-space path that all the manipulators take as $\bm{q}(s)$, which is a function of the scalar path coordinate $s$. The joint velocities and joint accelerations can then be obtained as $\dot{\bm{q}}(s) = \frac{d\bm{q}(s)}{ds}\frac{ds}{dt} = \bm{q}'(s)\dot{s}$ and $\Ddot{\bm{q}}(s) =
    \bm{q}''(s)\dot{s}^2 + \bm{q}'(s)\Ddot{s}$.
where, $\dot{s} = {ds}/{dt}$, $\Ddot{s} = {d^2s}/{dt^2}$, $\bm{q}'(s) = {d\bm{q}(s)}/{ds}$, and $\bm{q}''(s) = {d^2\bm{q}(s)}/{ds^2}$.

The equations of motion of the object can also be described using the scalar path coordinate $s$ as
\begin{align}
    \label{eq:object_eom_compact_in_s}
    \mathbf{G}_E(s)\bm{F}_E(s) + \mathbf{G}_M\bm{F}_M(s) + {^O\bm{f}_{ext}}(s) = {^{O}\bm{f}_{net}}(s).
\end{align}
Please note that, while the environment contact matrix $\mathbf{G}_E(s)$ may change depending upon the motion of the object. However, the manipulator contact matrix $\mathbf{G}_M$ remains constant as long as contact constraints at the manipulator-object contacts are satisfied. Therefore by incorporating the friction cone constraints at the manipulator-object contact, the object can be considered as an extension of the manipulator's end-effector. This assumption lets us express the object's position, velocity and acceleration expressed in its reference frame ${O}$ by relating it to the $j$-th manipulator's end-effector velocity and acceleration as,
\begin{align}
    \label{eq:obj_pos_kin}
    {^O\mathbf{X}}(s) = 
    \mathbf{X}_j{^{M_j}}\mathbf{T}_{O},
\end{align}
\begin{align}
    \label{eq:obj_vel_kin}
    {^O\mathbf{V}}(s) = 
    \mathbf{J}^b_j(\bm{q}_j(s))\dot{\bm{q}}_j(s) = \mathbf{J}^b_j(\bm{q}_j(s)){\bm{q}'}(s)\dot{s},
\end{align}
\begin{align}
    \label{eq:obj_acc_kin_in_t}
    {^O\dot{\mathbf{V}}}(s)  = 
    \frac{d(\mathbf{J}^b_j)}{dt}\dot{\bm{q}}_j(s) + \mathbf{J}^b_j(\bm{q}_j(s))\ddot{\bm{q}}_j(s),
\end{align}
\begin{align}
    \label{eq:obj_acc_kin}
    \begin{split}
    {^O\dot{\mathbf{V}}}(s) = &\big({\mathbf{J}^b_j}'(s)
    \bm{q}'_j(s) + \mathbf{J}^b_j(\bm{q}_j(s))\bm{q}''_j(s)\big)\dot{s}^2 \\
    & \quad + \mathbf{J}^b_j(\bm{q}_j(s))\bm{q}'_j(s)\ddot{s},
    \end{split}
\end{align}
where, ${^{M_j}}\mathbf{T}_{O}$ is the relative rigid body transformation expressing the pose of the object frame ${O}$ in the manipulator's end-effector frame and,
\begin{align}
    \label{eq:jac_derivative}
    {\mathbf{J}^b_j}'(s) = \sum_{i=1}^{n_j}\frac{d\Big(\mathbf{J}^b_j(\bm{q}_j(s))\Big)}{dq_{j,i}}\frac{dq_{j,i}}{ds}.
\end{align}
We note here that the relative rigid body transformation between the $j$-th manipulator's end-effector reference frame $\{X_j\}$ and the object reference frame $\{O\}$ needs to be taken into account when computing the body Jacobian $\mathbf{J}^b_j$ for this to be true. Hereafter, we will use the $1$-st manipulator to determine ${^O\mathbf{V}}$ and ${^O\dot{\mathbf{V}}}$. However any one of the $v$ manipulators can be used for this purpose.
\begin{gather}
    \label{eq:obj_kin_manip_1}
    {^O\mathbf{V}}(s) = \mathcal{J}_O(s)\dot{s},
    \quad
    {^O\dot{\mathbf{V}}}(s) = \mathcal{J}_O'(s)\dot{s}^2 + \mathcal{J}_O(s)\ddot{s},
    \\
    \mathcal{J}_O(s) = \mathbf{J}^b_1(\bm{q}_1(s))\bm{q}'_1(s),
    \\
    \mathcal{J}_O'(s) = {\mathbf{J}^b_1}'(s)
    \bm{q}'_1(s) + \mathbf{J}^b_1(\bm{q}_1(s))\bm{q}''_1(s),
\end{gather}

The net wrench acting on the object whose mass is $m$ kg and mass moment of inertia is $^O\mathcal{I}$ can then be expressed as
\begin{align}
    \label{eq:obj_net_wrench}
    {^{O}\bm{f}_{net}}(s) &= 
    \mathcal{M}_O{^O\dot{\mathbf{V}}}(s) +
    \mathcal{C}_O(s)\dot{s}^2,\\
    {^{O}\bm{f}_{net}}(s) &= \mathcal{M}_O\mathcal{J}_O(s)\ddot{s} +
    \left(\mathcal{M}_O\mathcal{J}'_O(s) + \mathcal{C}_O(s)\right)\dot{s}^2,
\end{align}
Here, $\mathcal{M}_O = 
\begin{bmatrix}
    \text{diag}(m) & \mathbf{0} \\
    \mathbf{0} & {^O\mathcal{I}}
\end{bmatrix}$ is the object mass matrix and is a constant in the object reference frame and 
$\mathcal{C}_O$ is the Coriolis and centrifugal term of the object which is given by
\begin{align}
    \label{eq:obj_corr_term}
    \mathcal{C}_O(s)\dot{s}^2 &= 
    \begin{bmatrix}
        \bm{\omega}^b(s) \times m\bm{v}^b(s) \\
        \bm{\omega}^b(s) \times {^O\mathcal{I}}\bm{\omega}^b(s)
    \end{bmatrix},\\
    \mathcal{C}_O(s)\dot{s}^2 &= 
    \begin{bmatrix}
        \mathbf{S}(\mathcal{J}_{O_\omega}(s)) & \mathbf{0} \\
        \mathbf{0} & \mathbf{S}(\mathcal{J}_{O_\omega}(s))
    \end{bmatrix}
    \mathcal{M}_{O}
    \mathcal{J}_O(s)\dot{s}^2,
\end{align}

The friction cone constraints can also be described using the scalar path coordinate $s$ as
\begin{align}
    \label{eq:env_obj_friction_cone_in_s}
    \bm{f}_{E_i}(s) \in \mathcal{K}_{E_i}, i=1,2,\dots,u
\end{align}
\begin{align}
    \label{eq:manip_obj_friction_cone_in_s}
    \bm{f}_{M_j}(s) \in \mathcal{K}_{M_j}, j=1,2,\dots,v
\end{align}
The upper bound on the normal contact forces can be described using the scalar path coordinates $s$ as
\begin{align}
    \label{eq:max_normal_contact_force_constraint_in_s}
    \bm{F}_z(s) \leq \overline{\bm{F}}_z(s)
\end{align}

Substituting the expressions for $\dot{\bm{q}}(s)$ and $\ddot{\bm{q}}(s)$ in (\ref{eq:eom_of_m_manipulators}), we get the following
\begin{align}
    \label{eq:eom_of_m_manipulators_in_s}
    \bm{\tau}(s) + \mathcal{J}_M(s)^T\bm{F}_M(s) =
    \mathcal{M}_M(s)\Ddot{s} + \mathcal{C}_M(s)\dot{s}^2 + \mathcal{G}_M(s),
\end{align}
where
\begin{gather}
    \label{eq:manips_mass_matrix_in_s}
    \mathcal{M}_M(s) = \mathbf{M}(\bm{q}(s))\bm{q}'(s),
    \quad
    \mathcal{G}_M(s) = \mathbf{G}(\bm{q}(s)),
\end{gather}
\begin{align}
    \label{eq:manips_c_term_in_s}
    \mathcal{C}_M(s) = \mathbf{M}(\bm{q}(s))\bm{q}''(s) +
    \mathbf{C}(\bm{q}(s), \bm{q}'(s))\bm{q}'(s).
\end{align}

To also ensure that the joint actuators of the manipulators are within the actuator limits and not saturated, the following joint torque limit constraints need to be considered
\begin{align}
    \label{eq:joint_torque_constraints_in_s}
    \underline{\bm{\tau}} \leq \bm{\tau}(s) \leq \overline{\bm{\tau}}
\end{align}
where $\underline{\bm{\tau}}$ and $\overline{\bm{\tau}}$ are the lower and upper joint torque limits of the manipulators respectively.

If, the equations (\ref{eq:object_eom_compact_in_s}), (\ref{eq:env_obj_friction_cone_in_s}), (\ref{eq:manip_obj_friction_cone_in_s}),  (\ref{eq:max_normal_contact_force_constraint_in_s}), (\ref{eq:eom_of_m_manipulators_in_s}), (\ref{eq:joint_torque_constraints_in_s}) are satisfied simultaneously throughout the motion, then the object can be manipulated while satisfying the contact constraints.



\subsection{Time Optimal Control}

The extension of the problem of time-optimal control of a manipulator along a path, as in \cite{verscheure2009trajopt}, applied to $v$ manipulators with the added contacts can be expressed as,
\begin{gather}
    \label{eq:traj_opt}
    {\underset{T, s, \bm\tau, \bm{F}_E, \bm{F}_M}{\text{minimize}}}~T
\shortintertext{subject to (for $t \in [0, T]$),}
    \bm{\tau}(s) + \mathcal{J}_M(s)\bm{F}_M(s) =
    \mathcal{M}_M(s)\Ddot{s} + \mathcal{C}_M(s)\dot{s}^2 + \mathcal{G}_M(s)\notag\\
    \underline{\bm{\tau}} \leq \bm{\tau}(s) \leq \overline{\bm{\tau}}\notag\\
    \mathbf{G}_E(s)\bm{F}_E(s) + \mathbf{G}_M\bm{F}_M(s) + {^O\bm{f}_{ext}}(s) = {^{O}\bm{f}_{net}}(s)\notag\\
    {^{O}\bm{f}_{net}}(s) = \mathcal{M}_O\mathcal{J}_O(s)\ddot{s} +
    \left(\mathcal{M}_O\mathcal{J}'_O(s) + \mathcal{C}_O(s)\right)\dot{s}^2\notag\\
    \bm{f}_{E_i}(s) \in \mathcal{K}_{E_i}, i=1,2,\dots,u\notag\\
    \bm{f}_{M_j}(s) \in \mathcal{K}_{M_j}, j=1,2,\dots,v\notag\\
    \bm{F}_z(s) \leq \overline{\bm{F}}_z\notag\\
    s(0) = 0,
    s(T) = 1,
    \dot{s}(0) = \dot{s}_0,
    \dot{s}(T) = \dot{s}_T,
    \dot{s}(t) \geq 0\notag
\end{gather}

\subsection{Reformulation as a Convex Optimization Problem}
\begin{figure*}
\centering
    \includegraphics[width=\textwidth]{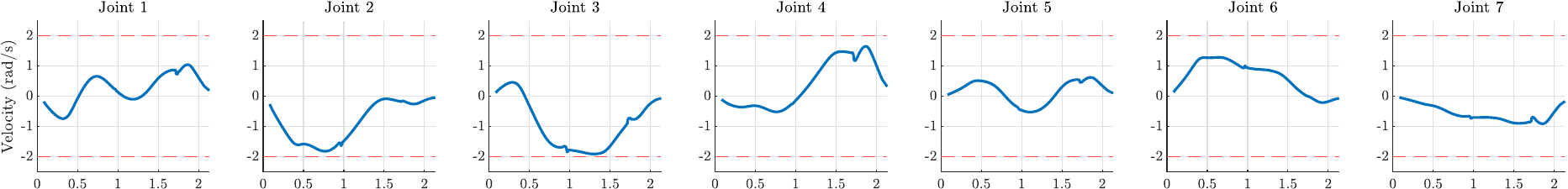}
    \includegraphics[width=\textwidth]{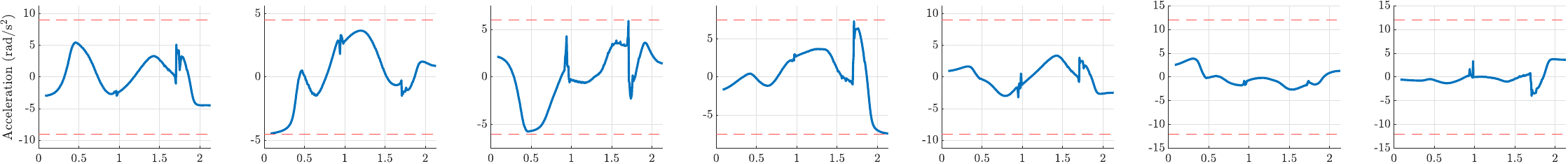}
    \includegraphics[width=\textwidth]{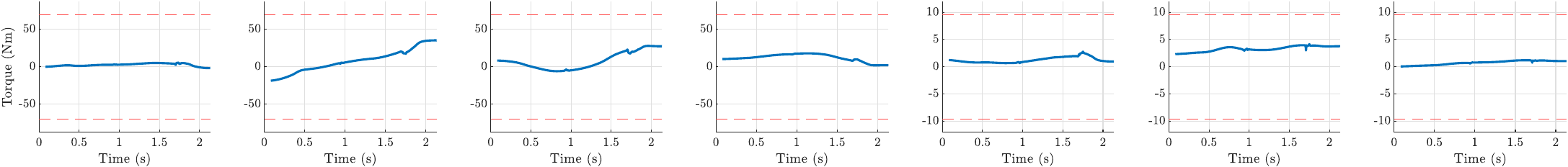}
    \caption{Plots of joint velocities, accelerations and torques for the task of picking up an object. The joint limits are shown using dashed red lines. Joints $3$ and $4$ are the ones getting closest to their velocity and acceleration limits.}
    \vspace{-15pt}
\label{fig:constrained_pick_and_place_plots}
\end{figure*}
We adopt the strategy of introduction of new optimization variables $a(s), b(s)$,as shown in \cite{verscheure2009trajopt}, to obtain a reformulated objective function as
\begin{align}
    \label{eq:objective_in_s}
    T = \int_{0}^{T}dt = \int_{0}^{1}\frac{1}{\dot{s}}ds
\end{align}
The new optimization variables are $a(s)=\ddot{s},b(s)=\dot{s}^2$, and by adding the constraint $b'(s) = 2a(s)$
results in,
\begin{align}
    \label{eq:change_of_var_constraint_derivation}
    \dot{b}(s) = 2\dot{s}\ddot{s} = 2\dot{s}a(s)
\end{align}
Thus, the optimization problem now becomes
\begin{gather}
    \label{eq:traj_opt_convex}
    {\underset{a , b,\bm\tau, \bm{F}_E, \bm{F}_M}{\text{minimize}}}~\int_{0}^{1}\frac{1}{\sqrt{b(s)}}ds\\
\shortintertext{subject to (for $s \in [0, 1]$),}
    \bm{\tau}(s) + \mathcal{J}_M(s)\bm{F}_M(s) = \mathcal{M}_M(s)a(s) + \mathcal{C}_M(s)b(s) + \mathcal{G}_M(s)\notag\\
    \underline{\bm{\tau}} \leq \bm{\tau}(s) \leq \overline{\bm{\tau}}\notag\\
    \mathbf{G}_E(s)\bm{F}_E(s) + \mathbf{G}_M\bm{F}_M(s) + {^O\bm{f}_{ext}}(s) = {^{O}\bm{f}_{net}}(s)\notag\\
    {^{O}\bm{f}_{net}}(s) = \mathcal{M}_O\mathcal{J}_O(s)a(s) +
    \left(\mathcal{M}_O{\mathcal{J}'_O}(s) + \mathcal{C}_O(s)\right)b(s)\notag\\
    \bm{f}_{E_i}(s) \in \mathcal{K}_{E_i}, i=1,2,\dots,u\notag\\
    \bm{f}_{M_j}(s) \in \mathcal{K}_{M_j}, j=1,2,\dots,v\notag\\
    \bm{F}_z(s) \leq \overline{\bm{F}}_z\notag\\
    b(0) = (\dot{s}(0))^2,
    b(T) = (\dot{s}(T))^2,
    b(s) \geq 0,
    b'(s) = 2a(s)\notag
\end{gather}
Here, the objective function is convex and all the constraints except for the SOC constraints $\mathcal{K}_{E_i}$ and $\mathcal{K}_{M_j}$ are linear functions of the optimization variables. Thus, the above optimization problem is a convex optimization problem.

\subsection{Manipulator Motion Constraints}
Apart from the actuator force/torque limits there can be other constraints such as velocity and acceleration limits in the joint space as well as in the task space which are convex constraints and can be included in the optimization. However, here we focus only on the joint space constraints.

\noindent\underline{Joint velocity constraints}: 
Linear inequality constraints
\begin{gather}
    \label{eq:joint_vel_constraints}
    0 \leq
    \dot{q}^2_{j,l}(s) \leq
    \overline{\dot{q}}^2_{j,l}\\
    0 \leq
    {q'_{j,l}}^2(s)\dot{s}^2 \leq
    \overline{\dot{q}}^2_{j,l}\\
    0 \leq
    {q'_{j,l}}^2(s)b(s) \leq
    \overline{\dot{q}}^2_{j,l}
\end{gather}
for all $j=1,\dots,v$ and $l=1,\dots,n_j$

\noindent\underline{Joint acceleration constraints}: Linear inequality constraints
\begin{gather}
    \label{eq:joint_acc_constraints}
    \underline{\ddot{\bm{q}}} \leq
    \ddot{\bm{q}}(s) \leq
    \overline{\ddot{\bm{q}}}\\
    \underline{\ddot{\bm{q}}} \leq
    \bm{q}''(s)\dot{s}^2 + \bm{q}'(s)\Ddot{s} \leq
    \overline{\ddot{\bm{q}}}\\
    \underline{\ddot{\bm{q}}} \leq
    \bm{q}''(s)b(s) + \bm{q}'(s)a(s) \leq
    \overline{\ddot{\bm{q}}}
\end{gather}




\section{Implementation And Solution Approach}
In this section, we detail the solution strategy for the optimization problem formulated in the previous section.

\subsection{Direct Transcription}
We follow the collocation approach proposed in \cite{verscheure2009trajopt} where $a(s)$ is assumed to be the piece-wise constant control input. This implies that $b(s)$ is piecewise linear and $\tau(s)$ is piecewise nonlinear. Thus, based on the values at the grid points, $b(s)$ becomes
\begin{align*}
    b(s) = b^k + \bigg(\frac{b^{k+1} - b^k}{s^{k+1} - s^k}\bigg)(s-s^k)\text{ for }s \in [s^k,s^{k+1}]
\end{align*}
Let us define $a^{k_m}, \tau^{k_m}, \bm{F}^{k_m}_E, \bm{F}^{k_m}_M$ be the values of the functions $a(s), \tau(s), \bm{F}^k_E(s), \bm{F}^k_M(s)$ at $s^{k_m} = \frac{s^{k+1}+s^{k}}{2}$, which is the midpoint of the grid points $s^k$ and $s^{k+1}$.


The objective function can now be expressed as
\begin{align}
    \int_{0}^{1}\frac{1}{\sqrt{b(s)}}ds &= 
    \sum_{k=0}^{K-1}\int_{s^k}^{s^{k+1}}\frac{1}{\sqrt{b(s)}}ds
    =\sum_{k=0}^{K-1}\frac{2(\Delta s^k)}{\sqrt{b^{k+1}} + \sqrt{b^k}}
\end{align}
where $\Delta s^k = s^{k+1}-s^k$.


However, this problem can be reformulated as a Second Order Cone Program (SOCP) and it is advantageous to do so as there are very effective methods for solving them \cite{boyd2007contact}.

\subsection{Second Order Cone Program}
The above direct transcription problem can be reformulated as a SOCP by introducing variables $c^k$ and $d^k$ with the addition of the following inequality constraints \cite{verscheure2009trajopt}
\begin{gather}
    c^k \leq \sqrt{b^k}, k = 0,\dots,K\\
    \frac{1}{c^{k+1} + c^k} \leq d^k, k = 0,\dots,K-1
\end{gather}
These constraints can also be expressed as SOC constraints
\begin{gather}
    \begin{Vmatrix}
        2c^k \\ b^k-1
    \end{Vmatrix}
    \leq b^k+1\\
    \begin{Vmatrix}
        2 \\ c^{k+1}+c^k-d^k
    \end{Vmatrix}
    \leq c^{k+1}+c^k+d^k
\end{gather}
This change in variable along with the introduction of the SOC constraints results in the following SOCP
\begin{gather}
    \label{eq:traj_opt_socp}
    {\underset{a^k , b^k, c^k, d^k, \bm{\tau}^{k_m}, \bm{F}^{k_m}_E, \bm{F}^{k_m}_M}{\text{minimize}}}~\sum_{k=0}^{K-1}{2(\Delta s^k)d^k}
\shortintertext{subject to,}
    \bm{\tau}^{k_m} + {\mathcal{J}_M^{k_m}}\bm{F}^{k_m}_M = \mathcal{M}^{k_m}_Ma^{k_m} + \mathcal{C}^{k_m}b^{k_m} + \mathcal{G}^{k_m}\notag\\
    \underline{\bm{\tau}} \leq \bm{\tau}^{k_m} \leq \overline{\bm{\tau}}\notag\\
    0 \leq (q'^{k_m}_{j,l})^2b^{k_m} \leq
    \overline{\dot{q}}_{j,l}^2~\forall~
    j=1,\dots,v\text{ and }l=1,\dots,n_j\notag\\
    \underline{\ddot{\bm{q}}} \leq
    \bm{q}''^{k_m}b^{k_m} + \bm{q}'^{k_m}a^{k_m} \leq
    \overline{\ddot{\bm{q}}}\notag\\
    \mathbf{G}^{k_m}_E\bm{F}^{k_m}_E + \mathbf{G}_M\bm{F}^{k_m}_M + {^O\bm{f}^{k_m}_{ext}} = {^{O}\bm{f}^{k_m}_{net}}\notag\\
    {^{O}\bm{f}^{k_m}_{net}}(s) = \mathcal{M}_O\mathcal{J}^{k_m}_Oa^{k_m} +
    \left(\mathcal{M}_O{\mathcal{J}'_O}^{k_m} + \mathcal{C}^{k_m}_O\right)b^{k_m}\notag\\
    \bm{f}^{k_m}_{E_i} \in \mathcal{K}_{E_i}, i=1,2,\dots,u\notag,
    \bm{f}^{k_m}_{M_j} \in \mathcal{K}_{M_j}, j=1,2,\dots,v\notag\\
    \bm{F}^{k_m}_z \leq \overline{\bm{F}}_z\notag,
    \begin{Vmatrix}
        2 \\ c^{k+1}+c^k-d^k
    \end{Vmatrix}
    \leq c^{k+1}+c^k+d^k\notag 
\intertext{for $k = 0 \dots K-1$,}
    \begin{Vmatrix}
        2c^k \\ b^k-1
    \end{Vmatrix}
    \leq b^k+1\notag,
    b^k \geq 0,~k = 0 \dots, K\notag\\
    b^0 = (\dot{s}^0)^2\notag,
    b^K = (\dot{s}^K)^2 \notag
\end{gather}
If $n=\sum_{j=1}^{v}{n_j}$ is the sum of the DOFs of all the $v$ manipulators, then for a choice of $K$ grid points, the total number of optimization variables is $K(4 + 3u + 4v + n) - 2$.






\section{Simulation Studies}\label{sec:sim_study}
\begin{figure*}
\centering
    \includegraphics[width=\textwidth]{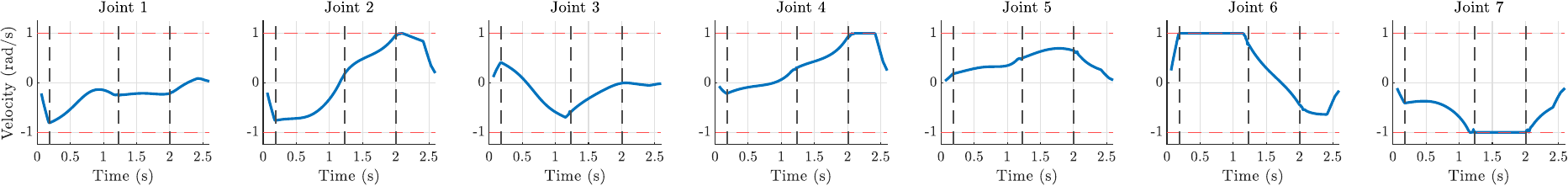}
    \vspace{-10pt}
    \caption{Plots of joint velocities for the pivoting task. The vertical dashed lines denote specific instances when the joints are at their limits and they correspond to the vertical dashed lines shown in the $(s,\dot{s})$ phase plane in Fig~\ref{fig:cover_pic}.}
    \vspace{-10pt}
    \label{fig:pivoting_joint_velocities}
\end{figure*}
\begin{figure*}
\centering
    \includegraphics[width=0.16\textwidth]{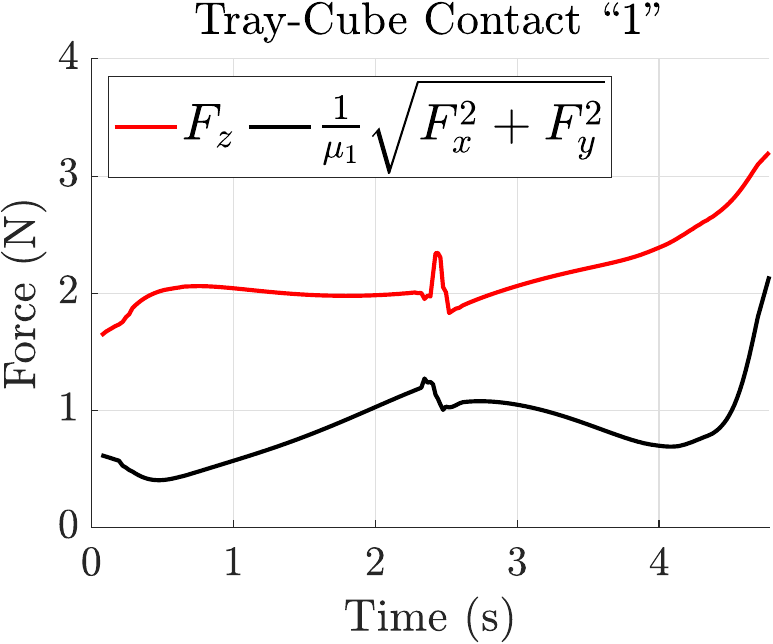}
    \includegraphics[width=0.16\textwidth]{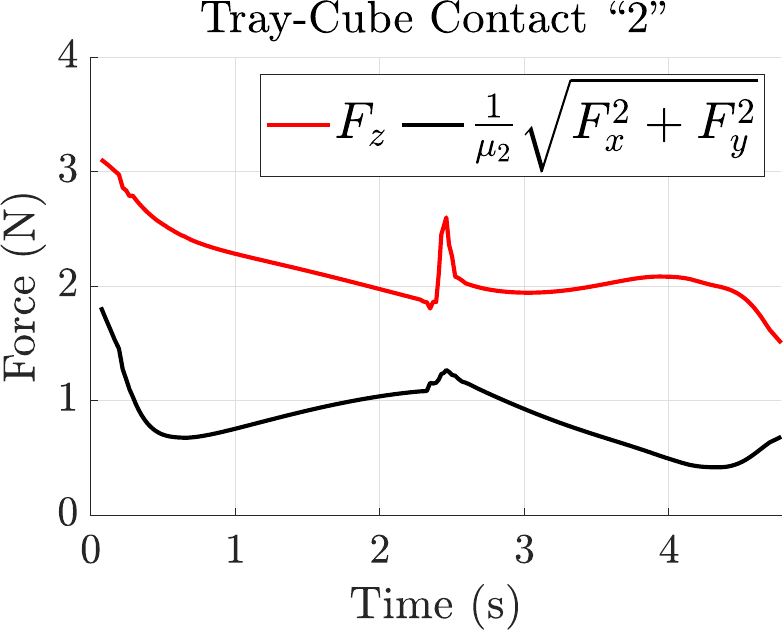}
    \includegraphics[width=0.16\textwidth]{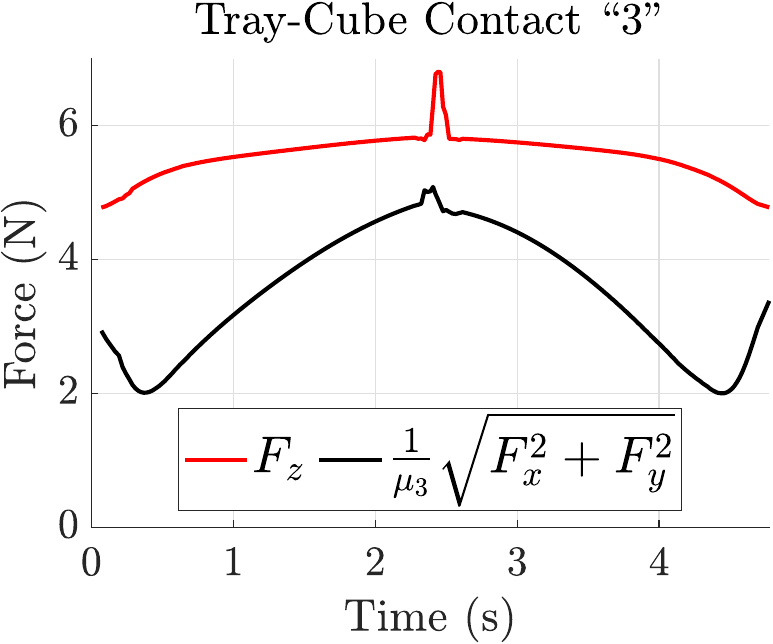}
    \includegraphics[width=0.16\textwidth]{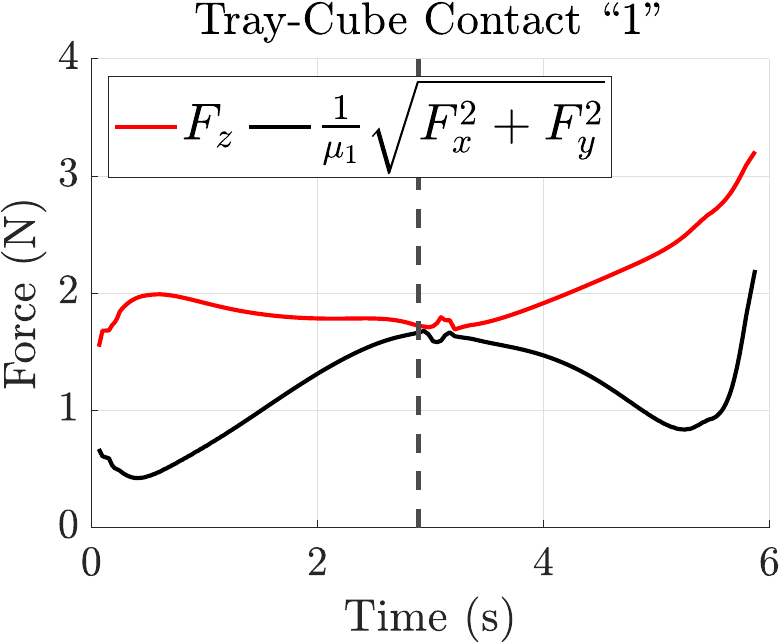}
    \includegraphics[width=0.16\textwidth]{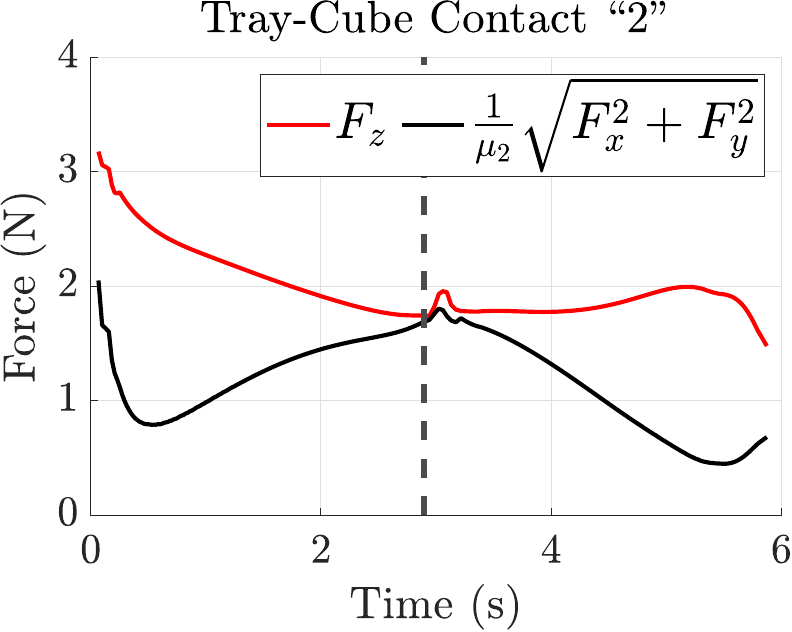}
    \includegraphics[width=0.16\textwidth]{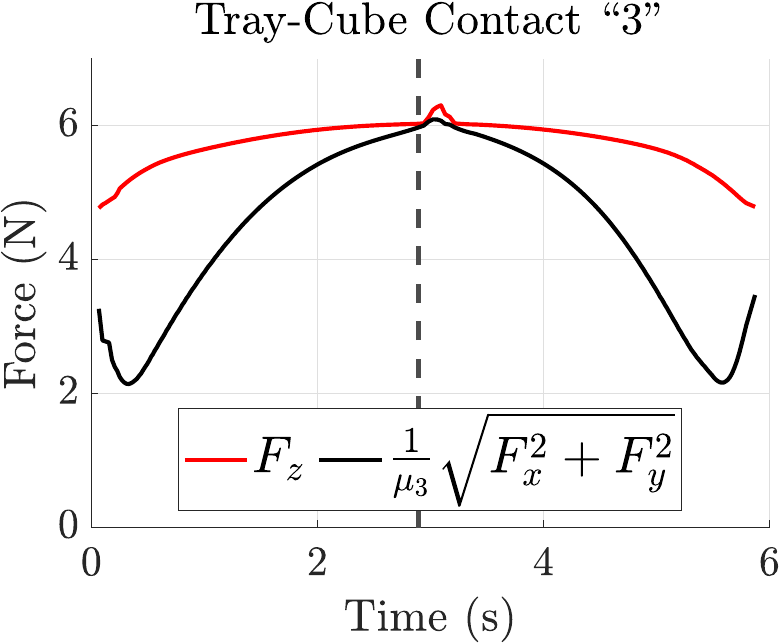}
    \caption{Tray-Cube Friction Cone Constraint results obtained for the non-prehensile task. First three images on the left show the constraints when the tray angle was set to $10^\circ$. The next three images show the constraints when the tray angle was set to $15^\circ$. The vertical dashed line in the last three images depict the constraints at their limits.}
    \vspace{-10pt}
\label{fig:tray_experiment_results}
\end{figure*}
We now present simulation experiments to evaluate our proposed formulation. We use YALMIP \cite{lofberg2004yalmip} which is a MATLAB toolbox for modeling and optimization with SeDuMi \cite{sturm1999sedumi} as the solver. All experiments were carried out on a laptop running Ubuntu $20.04$, equipped with an Intel Core i$9-9880$H CPU having $16$ cores. The $7$ DOF Frank Emika Panda research robot with a two finger parallel jaw gripper (see Figures \ref{fig:cover_pic} and \ref{fig:experiment_overview}) is used as the experimental test-bed. For all experiments, we set the torque bounds of the manipulator to be $80\%$ of the manipulator's rated torque limits\footnote{$\bm{F}_z$ was set to be less than or equal to the maximum continuous grasping force of the Franka Hand gripper.} and acceleration bounds to be $60\%$ of the manipulator's rated acceleration limits (see datasheet~\cite{panda_specifications} for robot hardware limits).
We use a Screw Linear Interpolation (ScLERP) based motion planner with Jacobian pseudoinverse \cite{sarker2020sclerp} to determine the joint space motion plan that the robot needs to execute.
A key fact to note is that the optimization problem does not modify the joint space path. It only solves for the time-parameterization of the path parameter $s(t)$.

We use $3$ exemplar tasks for validation of our approach: 
(a) pivoting an object while maintaining continuous contact with environment, (b) picking up an object from a flat surface, and (c) a non-prehensile manipulation task where the robot manipulates an object on a tray, which is called the `General Waiter Motion Problem'. We demonstrate that if there exists a minimum execution time solution that can satisfy the dynamic constraints, then by solving this optimization problem, we can determine the time parameterization of the path parameter $s(t)$, joint velocities $\dot{\bm{q}}(s(t))$ joint accelerations $\Ddot{\bm{q}}(s(t))$, torques $\bm{\tau}(s(t))$, manipulator-object forces $\bm{F}_M(s(t))$ and object-environment contact forces $\bm{F}_E(s(t))$, such that the trajectory execution time is a minimum. This is due to the fact that the optimization problem is convex, and hence any local optimum is also the global optimum. (Videos of Experiments--- \url{https://youtu.be/MJtQPyDUVXs}). 




\subsection{Use-Case \#1: Time-Optimal Object Pivoting}

The objective of this task is to use a manipulator to successfully pivot an cuboidal object while maintaining continuous contact with the environment (Fig. \ref{fig:cover_pic}).
The mass of the cuboid was set as $2$ kg, the values of the friction parameters were chosen as $\mu_M=0.4,~e_{E_*} = e_{M_x} = e_{M_y} = 1,~e_{M_z}=0.25$, and the number of grid points was chosen as $K = 250$. $\bm{F}_z$ was set as $70$ N for all manipulator-object contacts and the joint space velocity bounds were set to $\pm1$ rad/s for all joints. We consider two manipulator-object contacts and two environment-object contacts at the vertices of the pivoting edge (line contact). Although the formulation given in (\ref{eq:traj_opt_socp}) considers only one contact per manipulator, it can also be generalized to $w$ manipulator-object contacts for the $j$-th manipulator by introducing a friction cone constraint (\ref{eq:sfce_constraint}) for each contact. We carried out multiple trials with values of $\mu_E$ $0.2, 0.3, 0.4$ and $0.5$. In all trials, the manipulator started at the same configuration and the object paths are the same.  The joint velocities obtained for the pivoting task with $\mu_E=0.4$ are shown in Fig~\ref{fig:pivoting_joint_velocities}. We can see that the joints $6, 7$ and $4$ reach the velocity limits during execution. Due to the velocity limits being reached in all the trials and identical paths for the object and manipulator, we observed the same optimal trajectory execution time of $2.5972$ seconds across all the trials. However, if we increase the mass of the object beyond a certain value (depending on the choice of $\mu_E$), we get infeasibility, meaning that the manipulator cannot pivot the object within the given actuator/contact force limits. Note that it takes approximately $2.5$ seconds to obtain the solution.

\subsection{Use-Case \#2: Minimum Time Object Pick-up}
\begin{figure*}[ht!]
    \centering
    \includegraphics[width=\textwidth]{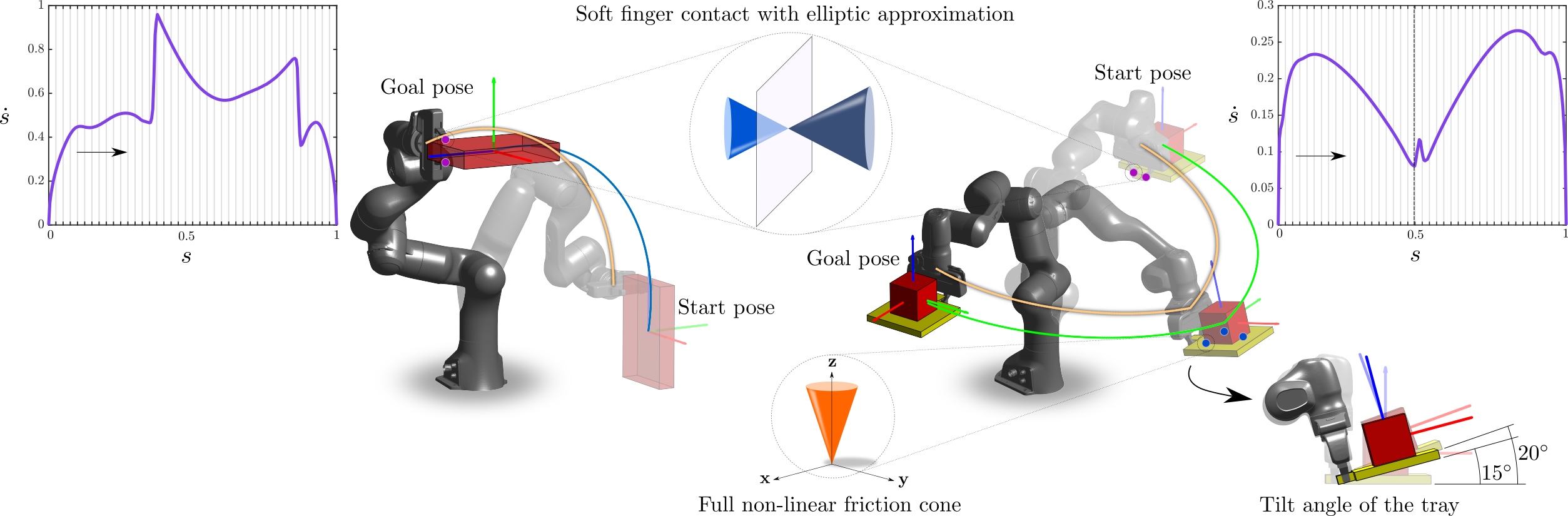}
    \caption{\textbf{Left}: Picking up an object along a constrained path in a minimum time. \textbf{Right}: Non-prehensile manipulation task: Waiter Motion Problem. The phase plane at the top left and right corners depict the limits of the system in the two use-cases as the scaling evolves.}
    \vspace{-20pt}
    \label{fig:experiment_overview}
\end{figure*}
We evaluate our proposed approach for the task of picking up a object (that is of cuboid shape) from a flat surface as shown in Fig \ref{fig:experiment_overview}. Note that for this task we do not consider any contact of the object with the environment. 
We only consider the two manipulator-object contacts (see Figure \ref{fig:experiment_overview}). The number of grid points was chosen as $K = 250$, the mass of the cuboid was set as $1$ kg and the values of the friction parameters were chosen to be $\mu_M=0.6,~e_{M_x} = e_{M_y} = 1,~e_{M_z}=0.25$.  $\bm{F}_z$ was set as $70$ N for all manipulator-object contacts and the joint space velocity bounds were set to $\pm2$ rad/s for all joints. We carried out multiple trials with the values for object mass $m(\text{kg})$ from $\{0.5, 0.75, 1.0, 1.25, 1.5, 1.75\}$. We observed an increase in the optimal trajectory execution time of $1.99$ seconds for $m=0.5\text{ kg}$ upto $3.2$ seconds for $m=1.25\text{ kg}$. Since we constrain the maximum magnitude of normal force that can be applied at the manipulator-object contacts, the problem becomes infeasible for $m \geq 1.5$ kg. The results shown in Fig~\ref{fig:constrained_pick_and_place_plots} are for $m=1$ kg. After modeling, YALMIP took approximately $1.5$ seconds for generating the output.


\subsection{Use-Case \#3: General Waiter Motion Problem}
We evaluate our approach for the general waiter motion problem which is an instance of non-prehensile manipulation. 
The goal of the task is to move the tray with the cube on top from the starting pose to the end pose on the desired trajectory as fast as possible (Fig~\ref{fig:experiment_overview}). 

Although this problem has been widely studied in the context of trajectory optimization~\cite{flores2013time,heins2023keep,muchacho2022solution,gattringer2023point}, none of the approaches takes into account contact constraints. Of the few that do~\cite{selvaggio2023non}, the usual norm is to approximate the friction cones with polygons for fast computation. Our approach, on the other hand, is general enough to take into account contact constraints between the cube and the tray as well (Eq. \ref{eq:traj_opt_socp}). Since we use a parallel jaw gripper to grasp the tray, we introduce two manipulator-tray contacts (SFCE model). The surface contact between the tray and the cube is modeled using three point contacts (PCWF model) (see Fig~\ref{fig:experiment_overview}).

Since we have two objects, we have two associated equations of motion. The equations of motion for the cube are
\begin{gather}
    {^{O_C}\bm{f}_{net}}(s) = \mathcal{M}_{O_C}\mathcal{J}_{O_C}(s)\ddot{s} +
    \left(\mathcal{M}_{O_C}\mathcal{J}'_{O_C}(s) + \mathcal{C}_{O_C}(s)\right)\dot{s}^2\notag\\
    \mathbf{G}_{CT}(s)\bm{F}_{CT}(s) + {^{O_C}\bm{f}_{C,ext}}(s) = {^{O_C}\bm{f}_{C,net}}(s)\notag
\end{gather}
and the equations of motion associated with the tray are
\begin{gather}
    {^{O_T}\bm{f}_{net}}(s) = \mathcal{M}_{O_T}\mathcal{J}_{O_T}(s)\ddot{s} +
    \left(\mathcal{M}_{O_T}\mathcal{J}'_{O_T}(s) + \mathcal{C}_{O_T}(s)\right)\dot{s}^2\notag\\
    \begin{split}
    \mathbf{G}_{CT}(s)(-\bm{F}_{CT}(s)) + \mathbf{G}_{M}(s)&(\bm{F}_{M}(s)) +
    {^{O_T}\bm{f}_{T,ext}}(s) \\
    &= {^{O_T}\bm{f}_{T,net}}(s)\notag
    \end{split}
\end{gather}
Where ${^{O_C}\bm{f}_{C,net}}\text{ and }{^{O_C}\bm{f}_{C,ext}}$ are the net wrench and the external wrench acting on the cube expressed in the reference frame of the cube $(\{O_c\})$ and ${^{O_T}\bm{f}_{T,net}}\text{ and }{^{O_T}\bm{f}_{T,ext}}$ are the net wrench and external wrench acting on the tray expressed in the reference frame of the tray $(\{O_T\})$. By adding these equality constraints and the friction cone constraints at all the contacts to the time optimal control (\ref{eq:traj_opt}), we can formulate the non-prehensile manipulation task as a SOCP. For this experiment, we modified the intermediate pose of the tray such that the angle the tray makes with the horizontal increases between different trials (Fig~\ref{fig:experiment_overview}). We use tilt angles of $0^\circ, 5^\circ, 10^\circ$ and $15^\circ$. The coefficient of friction at the three tray-cube contacts was set as $\mu_1=\mu_2=\mu_3=0.275$. The other parameters were set as $\mu_M = 0.4, e_{M_x}=e_{M_y}=1, e_{M_z}=0.25, F_z = 50\text{ N}, m_{tray}=0.125\text{ kg}, m_{cube}=1\text{ kg}, K = 250$. The plots in Fig~\ref{fig:tray_experiment_results} show the friction cone constraints at the tray-cube contacts when the tilt angle is $10^\circ$ (first three from left) and when the tilt angle is $15^\circ$. When the tilt angle, for the intermediate pose, was set to $16^\circ$, the optimization problem becomes infeasible, as the tray-cube contact constraints cannot be satisfied. This is evident from the fact that the contact constraints are at the limit for a tilt angle of $15^\circ$. We observed an optimal trajectory execution time of $4.14$ seconds for a $0^\circ$ tilt angle up to $5.87$ seconds for a $15^\circ$ tilt angle. For these trials, YALMIP took approximately $1.6$ seconds on average to get the output after it is given the modeled problem.

\section{Conclusion}\label{sec:conclusion}
In this paper we have shown that the problem of time optimal motion generation considering the entire dynamics of the system and the full nonlinear friction cone constraints at the contacts can be formulated as a SOCP. Our formulation is general enough to consider $v$ manipulators, each with one or more contacts with the manipulated object, as well as contacts of the object with the environment, if present. We also show that within the same framework, we can formulate non-prehensile manipulation tasks while satisfying contact constraints. The solution to our SOCP gives the time parameterization of the motion, joint torques of the manipulators, and also the contact forces at the manipulator-object and object-environment contacts. We evaluate our formulation on tasks like pivoting, where we have constraints on the motion through environmental contact, as well as tasks where we have constraints on the motion without environmental contact. 
In the future, we plan to study the effect of noise and model uncertainties on our proposed approach. We also aim to perform real-world experiments with multiple manipulators to investigate how the increase in computational load affects the time taken to solve the optimization problem.

\bibliographystyle{IEEEtran}
\bibliography{bibliography}

\begin{thebibliography}{10}
\providecommand{\url}[1]{#1}
\csname url@rmstyle\endcsname
\providecommand{\newblock}{\relax}
\providecommand{\bibinfo}[2]{#2}
\providecommand\BIBentrySTDinterwordspacing{\spaceskip=0pt\relax}
\providecommand\BIBentryALTinterwordstretchfactor{4}
\providecommand\BIBentryALTinterwordspacing{\spaceskip=\fontdimen2\font plus
\BIBentryALTinterwordstretchfactor\fontdimen3\font minus \fontdimen4\font\relax}
\providecommand\BIBforeignlanguage[2]{{%
\expandafter\ifx\csname l@#1\endcsname\relax
\typeout{** WARNING: IEEEtran.bst: No hyphenation pattern has been}%
\typeout{** loaded for the language `#1'. Using the pattern for}%
\typeout{** the default language instead.}%
\else
\language=\csname l@#1\endcsname
\fi
#2}}

\bibitem{zhou2022review}
C.~Zhou, B.~Huang, and P.~Fr{\"a}nti, ``A review of motion planning algorithms for intelligent robots,'' \emph{Journal of Intelligent Manufacturing}, vol.~33, no.~2, pp. 387--424, 2022.

\bibitem{laha2022coordinate}
R.~Laha, R.~Sun, W.~Wu, D.~Mahalingam, N.~Chakraborty, L.~F. Figueredo, and S.~Haddadin, ``Coordinate invariant user-guided constrained path planning with reactive rapidly expanding plane-oriented escaping trees,'' in \emph{2022 International Conference on Robotics and Automation (ICRA)}.\hskip 1em plus 0.5em minus 0.4em\relax IEEE, 2022, pp. 977--984.

\bibitem{laha2023predictive}
R.~Laha, M.~Becker, J.~Vorndamme, J.~Vrabel, L.~F. Figueredo, M.~A. M{\"u}ller, and S.~Haddadin, ``Predictive multi-agent based planning and landing controller for reactive dual-arm manipulation,'' \emph{IEEE Transactions on Robotics}, 2023.

\bibitem{lin1983formulation}
C.~Lin, P.~Chang, and J.~Luh, ``Formulation and optimization of cubic polynomial joint trajectories for industrial robots,'' \emph{IEEE Transactions on automatic control}, vol.~28, no.~12, pp. 1066--1074, 1983.

\bibitem{bobrow1985time}
J.~E. Bobrow, S.~Dubowsky, and J.~S. Gibson, ``Time-optimal control of robotic manipulators along specified paths,'' \emph{The international journal of robotics research}, vol.~4, no.~3, pp. 3--17, 1985.

\bibitem{shiller1989robot}
Z.~Shiller and S.~Dubowsky, ``Robot path planning with obstacles, actuator, gripper, and payload constraints,'' \emph{IJRR}, 1989.

\bibitem{suomalainen2022survey}
M.~Suomalainen, Y.~Karayiannidis, and V.~Kyrki, ``A survey of robot manipulation in contact,'' \emph{Robotics and Autonomous Systems}, 2022.

\bibitem{verscheure2008practical}
D.~Verscheure, B.~Demeulenaere, J.~Swevers, J.~De~Schutter, and M.~Diehl, ``Practical time-optimal trajectory planning for robots: a convex optimization approach,'' \emph{IEEE Transactions on Automatic Control}, vol.~14, p.~28, 2008.

\bibitem{laha2023s}
R.~Laha, W.~Wu, R.~Sun, N.~Mansfeld, L.~F. Figueredo, and S.~Haddadin, ``S*: On safe and time efficient robot motion planning,'' in \emph{2023 IEEE International Conference on Robotics and Automation (ICRA)}.\hskip 1em plus 0.5em minus 0.4em\relax IEEE, 2023, pp. 12\,758--12\,764.

\bibitem{kao2016contact}
I.~Kao, K.~M. Lynch, and J.~W. Burdick, ``Contact modeling and manipulation,'' \emph{Springer Handbook of Robotics}, pp. 931--954, 2016.

\bibitem{patankar2020handobject}
A.~Patankar, A.~Fakhari, and N.~Chakraborty, ``Hand-object contact force synthesis for manipulating objects by exploiting environment,'' in \emph{2020 IEEE/RSJ International Conference on Intelligent Robots and Systems (IROS)}, 2020, pp. 9182--9189.

\bibitem{fakhari2021motion}
A.~Fakhari, A.~Patankar, and N.~Chakraborty, ``Motion and force planning for manipulating heavy objects by pivoting,'' in \emph{2021 IEEE/RSJ Inter. Conf. on Intelligent Robots and Systems (IROS)}.

\bibitem{lynch2017modern}
K.~M. Lynch and F.~C. Park, \emph{Modern robotics}.\hskip 1em plus 0.5em minus 0.4em\relax Cambridge University Press, 2017.

\bibitem{selvaggio2023non}
M.~Selvaggio, A.~Garg, F.~Ruggiero, G.~Oriolo, and B.~Siciliano, ``Non-prehensile object transportation via model predictive non-sliding manipulation control,'' \emph{IEEE Trans. Control Syst. Technol.}, 2023.

\bibitem{kahn1969near}
M.~E. Kahn, \emph{The Near-minimum-time Control of Open-loop Articulated Kinematic Chains}.\hskip 1em plus 0.5em minus 0.4em\relax Stanford University, 1969, no. 106.

\bibitem{kahn1971near}
M.~Kahn, ``The near-minimum-time control of open-loop articulated kinematic chains,'' \emph{Trans. ASME, J. of Dyn. Sys. Meas. and Contr.}, vol.~93, pp. 164--172, 1971.

\bibitem{niv1984optimal}
M.~Niv and D.~M. Auslander, ``Optimal control of a robot with obstacles,'' in \emph{1984 American Control Conference}.\hskip 1em plus 0.5em minus 0.4em\relax IEEE, 1984.

\bibitem{hollerbach1983dynamic}
J.~M. Hollerbach, ``Dynamic scaling of manipulator trajectories,'' in \emph{1983 American Control Conference}.\hskip 1em plus 0.5em minus 0.4em\relax IEEE, 1983, pp. 752--756.

\bibitem{shin1985minimum}
K.~Shin and N.~McKay, ``Minimum-time control of robotic manipulators with geometric path constraints,'' \emph{IEEE Transactions on Automatic Control}, vol.~30, no.~6, pp. 531--541, 1985.

\bibitem{kunz2013time}
T.~Kunz and M.~Stilman, ``Time-optimal trajectory generation for path following with bounded acceleration and velocity,'' \emph{Robotics: Science and Systems VIII}, p. 209, 2013.

\bibitem{pham2014general}
Q.-C. Pham, ``A general, fast, and robust implementation of the time-optimal path parameterization algorithm,'' \emph{IEEE Transactions on Robotics}, vol.~30, no.~6, pp. 1533--1540, 2014.

\bibitem{lange2015path}
F.~Lange and A.~Albu-Sch{\"a}ffer, ``Path-accurate online trajectory generation for jerk-limited industrial robots,'' \emph{IEEE Robotics and Automation Letters}, vol.~1, no.~1, pp. 82--89, 2015.

\bibitem{verscheure2009trajopt}
D.~Verscheure, B.~Demeulenaere, J.~Swevers, J.~De~Schutter, and M.~Diehl, ``Time-optimal path tracking for robots: A convex optimization approach,'' \emph{IEEE Transactions on Automatic Control}, vol.~54, no.~10, pp. 2318--2327, 2009.

\bibitem{kroger2009online}
T.~Kr{\"o}ger and F.~M. Wahl, ``Online trajectory generation: Basic concepts for instantaneous reactions to unforeseen events,'' \emph{IEEE Transactions on Robotics}, vol.~26, no.~1, pp. 94--111, 2009.

\bibitem{he2022online}
S.~He, C.~Hu, S.~Lin, and Y.~Zhu, ``An online time-optimal trajectory planning method for constrained multi-axis trajectory with guaranteed feasibility,'' \emph{IEEE Robotics and Automation Letters}, vol.~7, no.~3, pp. 7375--7382, 2022.

\bibitem{lobo1998applications}
M.~S. Lobo, L.~Vandenberghe, S.~Boyd, and H.~Lebret, ``Applications of second-order cone programming,'' \emph{Linear algebra and its applications}, vol. 284, no. 1-3, pp. 193--228, 1998.

\bibitem{boyd2007contact}
S.~P. Boyd and B.~Wegbreit, ``Fast computation of optimal contact forces,'' \emph{IEEE Transactions on Robotics}, vol.~23, no.~6, pp. 1117--1132, 2007.

\bibitem{sleiman2019contact}
J.-P. Sleiman, J.~Carius, R.~Grandia, M.~Wermelinger, and M.~Hutter, ``Contact-implicit trajectory optimization for dynamic object manipulation,'' in \emph{2019 IEEE/RSJ International Conference on Intelligent Robots and Systems (IROS)}.\hskip 1em plus 0.5em minus 0.4em\relax IEEE, 2019, pp. 6814--6821.

\bibitem{onol2018comparative}
A.~O. Onol, P.~Long, and T.~Padlr, ``A comparative analysis of contact models in trajectory optimization for manipulation,'' in \emph{2018 IEEE/RSJ International Conference on Intelligent Robots and Systems (IROS)}.\hskip 1em plus 0.5em minus 0.4em\relax IEEE, 2018, pp. 1--9.

\bibitem{onol2019contact}
A.~{\"O}. {\"O}nol, P.~Long, and T.~Pad{\i}r, ``Contact-implicit trajectory optimization based on a variable smooth contact model and successive convexification,'' in \emph{2019 International Conference on Robotics and Automation (ICRA)}.\hskip 1em plus 0.5em minus 0.4em\relax IEEE, 2019, pp. 2447--2453.

\bibitem{onol2020tuning}
A.~{\"O}. {\"O}nol, R.~Corcodel, P.~Long, and T.~Pad{\i}r, ``Tuning-free contact-implicit trajectory optimization,'' in \emph{2020 IEEE International Conference on Robotics and Automation (ICRA)}.\hskip 1em plus 0.5em minus 0.4em\relax IEEE, 2020.

\bibitem{murray2017mathematical}
R.~M. Murray, Z.~Li, and S.~S. Sastry, \emph{A mathematical introduction to robotic manipulation}.\hskip 1em plus 0.5em minus 0.4em\relax CRC press, 2017.

\bibitem{boyd2004convex}
S.~P. Boyd and L.~Vandenberghe, \emph{Convex optimization}.\hskip 1em plus 0.5em minus 0.4em\relax Cambridge university press, 2004.

\bibitem{lofberg2004yalmip}
J.~L{\"{o}}fberg, ``Yalmip : A toolbox for modeling and optimization in matlab,'' in \emph{In Proceedings of the CACSD Conference}, Taipei, Taiwan.

\bibitem{sturm1999sedumi}
\BIBentryALTinterwordspacing
J.~F. Sturm, ``Using sedumi 1.02, a {MATLAB} toolbox for optimization over symmetric cones,'' \emph{Optimization Methods and Software}, vol.~11, no. 1-4, pp. 625--653, 1999. [Online]. Available: \url{https://doi.org/10.1080/10556789908805766}
\BIBentrySTDinterwordspacing

\bibitem{panda_specifications}
F.~Emika, ``Franka emika panda hardware specifications,'' \url{https://frankaemika.github.io/docs/control_parameters.html#limits-for-panda}, accessed: March 2024.

\bibitem{sarker2020sclerp}
A.~Sarker, A.~Sinha, and N.~Chakraborty, ``On screw linear interpolation for point-to-point path planning,'' in \emph{2020 IEEE/RSJ International Conference on Intelligent Robots and Systems (IROS)}, 2020.

\bibitem{flores2013time}
F.~G. Flores and A.~Kecskem{\'e}thy, ``Time-optimal path planning for the general waiter motion problem,'' in \emph{Advances in Mechanisms, Robotics and Design Education and Research}.\hskip 1em plus 0.5em minus 0.4em\relax Springer, 2013, pp. 189--203.

\bibitem{heins2023keep}
A.~Heins and A.~P. Schoellig, ``Keep it upright: Model predictive control for nonprehensile object transportation with obstacle avoidance on a mobile manipulator,'' \emph{IEEE Robot. Autom. Lett.}, 2023.

\bibitem{muchacho2022solution}
R.~I.~C. Muchacho, R.~Laha, L.~F. Figueredo, and S.~Haddadin, ``A solution to slosh-free robot trajectory optimization,'' in \emph{2022 IEEE/RSJ International Conference on Intelligent Robots and Systems (IROS)}.\hskip 1em plus 0.5em minus 0.4em\relax IEEE, 2022, pp. 223--230.

\bibitem{gattringer2023point}
H.~Gattringer, A.~M{\"u}ller, S.~Weitzhofer, and M.~Sch{\"o}rgenhumer, ``Point to point time optimal handling of unmounted rigid objects and liquid-filled containers,'' \emph{Mechanism and Machine Theory}, vol. 184, p. 105286, 2023.

\end{thebibliography}

\end{document}